\documentclass[10pt,twocolumn,letterpaper]{article}

\usepackage[pagenumbers]{iccv} 

\definecolor{iccvblue}{rgb}{0.21,0.49,0.74}
\usepackage[pagebackref,breaklinks,colorlinks,allcolors=iccvblue]{hyperref}
\usepackage{caption}
\usepackage{subcaption}
\usepackage{graphicx}   
\usepackage{multirow} 
\newcommand{\modelname}{InteractAvatar\xspace}

\def\paperID{7884} 
\def\confName{ICCV}
\def\confYear{2025}

\title{InteractAvatar: Modeling Hand-Face Interaction in Photorealistic Avatars with Deformable Gaussians}

\author{Kefan Chen\textsuperscript{1}\textsuperscript{2}
\and
Sreyas Mohan\textsuperscript{*}\textsuperscript{2}
\and
Justin Theiss\textsuperscript{*}\textsuperscript{2}
\and
Sergiu Oprea\textsuperscript{*}\textsuperscript{2}
\and
Srinath Sridhar\textsuperscript{1}
\and
Aayush Prakash\textsuperscript{2}
\\ 
\centerline{%
\textsuperscript{1}Brown University
\hspace{2em} 
\textsuperscript{2}Meta Reality Labs}
}

\begin{document}

\twocolumn[{
    \maketitle
    \centering
    \begin{minipage}{\textwidth}
        \begin{minipage}[b]{0.288\textwidth}
            \centering
            \includegraphics[width=\textwidth]{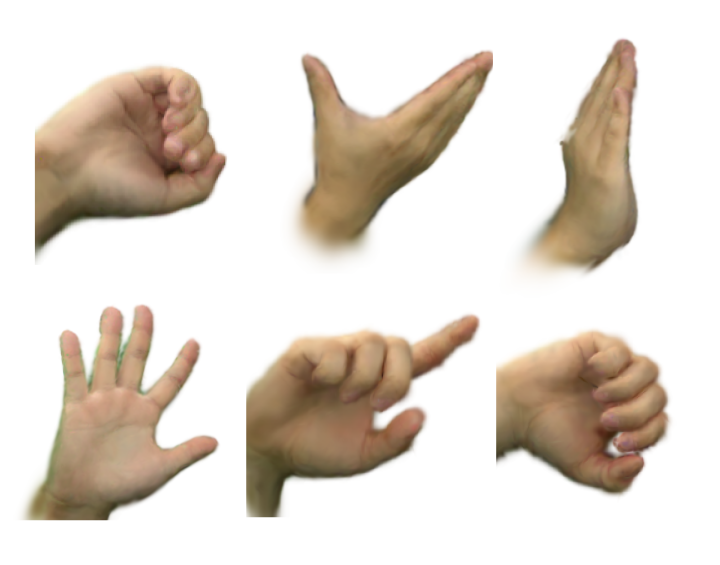}\\
            (a) Dynamic Gaussian Hand
            \label{fig:teaser_hand}
        \end{minipage}%
        \hfill
        \begin{minipage}[b]{0.404\textwidth}
            \centering
            \includegraphics[width=\textwidth]{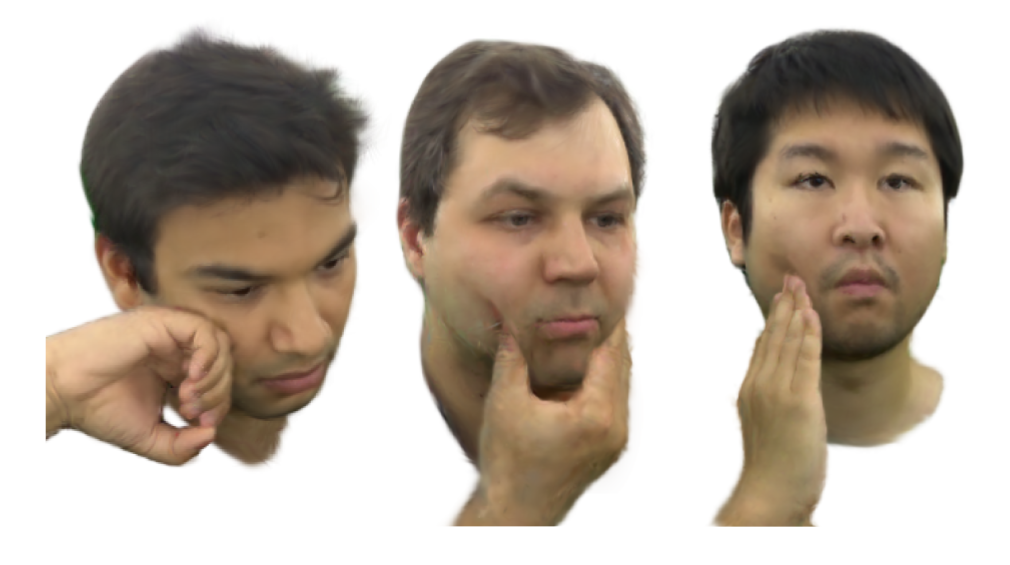}\\
            (b) Non-rigid hand-face interaction.
            \label{fig:teaser_interact}
        \end{minipage}%
        \hfill
        \begin{minipage}[b]{0.308\textwidth}
            \centering
            \includegraphics[width=\textwidth]{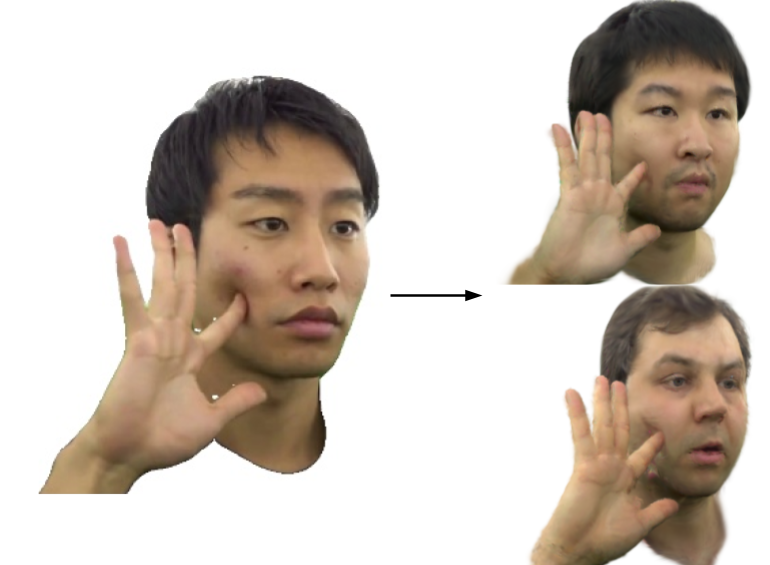}\\
            (c) Cross-actor enactment.
            \label{fig:teaser_enact}
        \end{minipage}
        \captionof{figure}{
        We propose \textbf{\modelname} which enables (a) \textbf{Dynamic Gaussian Hand}. Our novel representation anchors 3D Gaussian kernels to a hand template mesh and a learnable neural network allowing for pose-dependent articulation, self-cast shadows, and high-fidelity appearance modeling. (b) \textbf{Non-Rigid Hand-Face Interaction}. We introduce a learnable interaction module that refines hand-induced deformations and shading effects on the face, ensuring realistic skin contact dynamics. (c) \textbf{Cross-Actor Enactment}. We can transfer hand and face motions across different subjects, demonstrating its generalization capability to unseen identities and gestures.}
        \label{fig:teaser}
    \end{minipage}
    \vspace{5mm}
}]

\begin{abstract}

With increasing interest in digital avatars coupled with the importance of expressions and gestures in communication, modeling natural avatar behavior remains an important challenge in many industries such as teleconferencing, gaming, and AR/VR. Human hands are the primary tool for interacting with the environment and essential for realistic human behavior modeling, yet existing 3D hand and head avatar models often overlook the crucial aspect of hand-body interactions, such as between hand and face. We present \modelname, the first model to faithfully capture the photorealistic appearance of dynamic hand and non-rigid hand-face interactions. Our novel Dynamic Gaussian Hand model, combining template model and 3D Gaussian Splatting as well as a dynamic refinement module, captures pose-dependent change, e.g. the fine wrinkles and complex shadows that occur during articulation. Importantly, our hand-face interaction module models the subtle geometry and appearance dynamics that underlie common gestures.
Through experiments of novel view synthesis, self-reenactment, and cross-identity reenactment, we demonstrate that \modelname can reconstruct hand and hand-face interactions from monocular or multi-view videos with high-fidelity details and be animated with novel poses.
\end{abstract}    
\section{Introduction}
\label{sec:intro}
Human communication transcends mere words with expressions, gestures, and subtle physical interactions conveying instantly recognizable emotions, intentions, and empathy. Studies show that these nonverbal cues play a crucial role in interaction~\cite{asaliouglu2023role}, and failure to replicate them in digital avatars can disrupt comprehension and diminish immersion~\cite{adkins2023important, luo2012perceptual}.
Yet, faithfully capturing non-verbal communication modes in digital avatars remains a significant challenge.
In fields like teleconferencing~\cite{seymour2018actors, gunkel2018virtual, seymour2020facing}, VR/AR~\cite{seymour2020facing, pumarola2018ganimation, wei2019vr}, gaming~\cite{banks2016emotion, ratan2016mii}, and virtual social worlds~\cite{seymour2020facing}, there is a growing demand for avatars capable of dynamic, realistic interactions to enhance immersion and improve the quality of virtual human representation.


Although several methods exist to reconstruct photorealistic human avatars~\cite{INSTA:CVPR2023, mueller2022instant, zheng2022imavatar, hong2021headnerf, Zheng2023pointavatar, Gao2022nerfblendshape, xu2023avatarmav}, they largely ignore \emph{hand-face interactions} - an essential aspect of natural human behavior. Studies show that people touch their face up to 600~\cite{kwok2015face} or 800~\cite{spille2021stop} times a day. Furthermore, simple, rigid interactions appear unnatural (see Figures~\ref{fig:nvs},~\ref{fig:self_enactment}), underscoring the need for models that capture the subtleties of hand-face interactions.

Entertainment industries have traditionally relied on customized technologies like LightStage~\cite{debevec2002lighting, bebevec2002light,martinez2024codec,saito2024rgca,NEURIPS2024_9712b783} or physics-based simulations~\cite{wheatland2015state, noh1998survey}, such as finite element methods (FEM)~\cite{gourret1989simulation, basu19983d, guenter1989system} and position-based dynamics (PBD)~\cite{roussellet2018dynamic, bender2015position, chentanez2020cloth, sun2024physhand}, to model the complex dynamics of skin deformation and contact between hands and face. However, these methods often require extremely specialized hardware for data capture, extensive manual parameter tuning, are computationally expensive, and often struggle to generalize across new poses or expressions.
Data-driven approaches, including Neural Radiance Fields~\cite{mildenhall2020nerf} and 3D Gaussian Splatting~\cite{kerbl3Dgaussians}, are now state-of-the-art in modeling face and hand appearance\cite{INSTA:CVPR2023, mueller2022instant, zheng2022imavatar, hong2021headnerf, Zheng2023pointavatar, Gao2022nerfblendshape, xu2023avatarmav}, but do not yet handle hand-face interactions.

We introduce \textbf{\modelname}, the first method to leverage 3D Gaussian Splatting for modeling \textbf{non-rigid hand-face deformations and interactions}. Our approach builds upon GaussianAvatar~\cite{qian2024gaussianavatars} by incorporating a hybrid mesh-Gaussian representation for both hands and faces, enabling high-fidelity animation and novel view synthesis. Our key innovations (see Figure~\ref{fig:teaser}) include:
\begin{itemize}
    \item \textbf{Dynamic Gaussian Hands}. Capturing hand appearance and dynamics is challenging due to the complex geometry, varied textures, and intricate movements driven by numerous bones, joints, and muscles~\cite{wu2024diceendtoenddeformationcapture, DecafTOG2023}. We propose a \emph{novel hand representation that anchors Gaussian kernels to a template mesh} and learns to \emph{dynamically adjust geometry} (position, scale, rotation) and \emph{appearance} (color, opacity). This allows accurate modeling of articulation-dependent effects like self-cast shadows and wrinkles, improving generalization to unseen poses.
    \item \textbf{Photorealistic Hand-Face Interaction Module}. 
    While previous works have studied mesh recovery from hand-face interaction, our avatar model captures \emph{fine-grained changes in facial geometry and appearance caused by hand contact}. This enables lifelike rendering of shadows, skin deformations, and pose-dependent visual effects.
    \item \textbf{New state-of-the-art}. Through extensive experiments on novel view synthesis, self-reenactment, and cross-identity reenactment, we demonstrate \emph{qualitatively} and \emph{quantitatively} that \modelname achieves superior realism in modeling hand-face interactions compared to prior methods, \emph{paving the way for more immersive digital experiences in AR/VR, gaming, and telepresence}.
\end{itemize}

\section{Relevant Work}
\label{sec:relevant_work}

\textbf{Photorealistic Avatars}. Methods like Neural Radiance Fields (NeRF)~\cite{hong2021headnerf} and 3D Gaussian Splatting~\cite{kerbl3Dgaussians}, which capture high-quality renderings of static scenes have been used to create photorealistic avatars~\cite{INSTA:CVPR2023, mueller2022instant, zheng2022imavatar, hong2021headnerf, Zheng2023pointavatar, Gao2022nerfblendshape, xu2023avatarmav}. Similar to these models, \modelname builds on 3D Gaussians Splatting enabling high-fidelity reconstruction.

\noindent \textbf{Neural Hand Rendering}. 
Precise hand modeling is essential for realistic digital avatars, yet prior methods have key limitations. MANUS~\cite{pokhariya2024manus} uses articulated 3D Gaussian kernels for accurate hand-object contact but requires extensive multi-camera data and complex capture setups. Further, MANUS only focuses on interactions with rigid objects, unlike our task which requires much more intricate modeling. LiveHand~\cite{mundra2023livehand}, though capable of real-time, photorealistic rendering, adopts a NeRF-based implicit representation and does not efficiently model interactions between entities. LISA~\cite{corona2022lisa} offers a versatile model that combines shape, appearance, and animation, yet it lacks high-fidelity interaction modeling. In contrast, \modelname’s \emph{dynamic Gaussian hand} (a) attaches Gaussian kernel to template mesh obtained through MANO~\cite{MANO:SIGGRAPHASIA:2017}, and use (b) learnable modules that allow pose dependent deformations and changes, allows us to efficiently reason contact and deformation while generalizing to unseen poses during animation.

\noindent \textbf{Hybrid Avatar Representations}. Hybrid avatar representations leverage both mesh-based geometry and implicit models for realistic, adaptable avatars. SCARF~\cite{Feng2022scarf} and DELTA~\cite{Feng2023DELTA} separate body structure from features like hair and clothing, blending meshes with neural fields to control each component individually. GaussianAvatar~\cite{qian2023gaussianavatars} and SplattingAvatar~\cite{shao2024splattingavatar} embed Gaussian kernels on the FLAME~\cite{FLAME:SiggraphAsia2017} mesh, allowing them to deform smoothly with pose changes and capture high-detail appearances without predefined skinning weights. \modelname adopts this approach. However, unlike static-feature approaches like Ref.~\cite{bai2023learningpersonalizedhighquality,ho2023custom}, which attach features to fixed mesh vertices, our model enable dynamic, pose-responsive deformations achieving better geometry and appearance.

\noindent \textbf{Hand-Face Interaction}. Reconstructing 3D hand-face interactions from images is challenging due to self-occlusions, diverse spatial relationships between hands and face, complex deformations, and the ambiguity of the single-view setting. DICE~\cite{wu2024diceendtoenddeformationcapture} estimates the poses of hands and faces, contacts, and deformations simultaneously using a Transformer-based architecture. DECAF~\cite{DecafTOG2023}, introduces a global fitting optimization guided by contact and deformation estimation networks trained on studio-collected data with 3D annotations. \modelname uses a DECAF-like algorithm to resolve hand-face collisions and obtain coarse per-vertex 3D offsets, but refines these using a novel learnable module for precise facial geometry and appearance.  

\begin{figure*}[!ht]
    \centering
    \includegraphics[width=\textwidth]{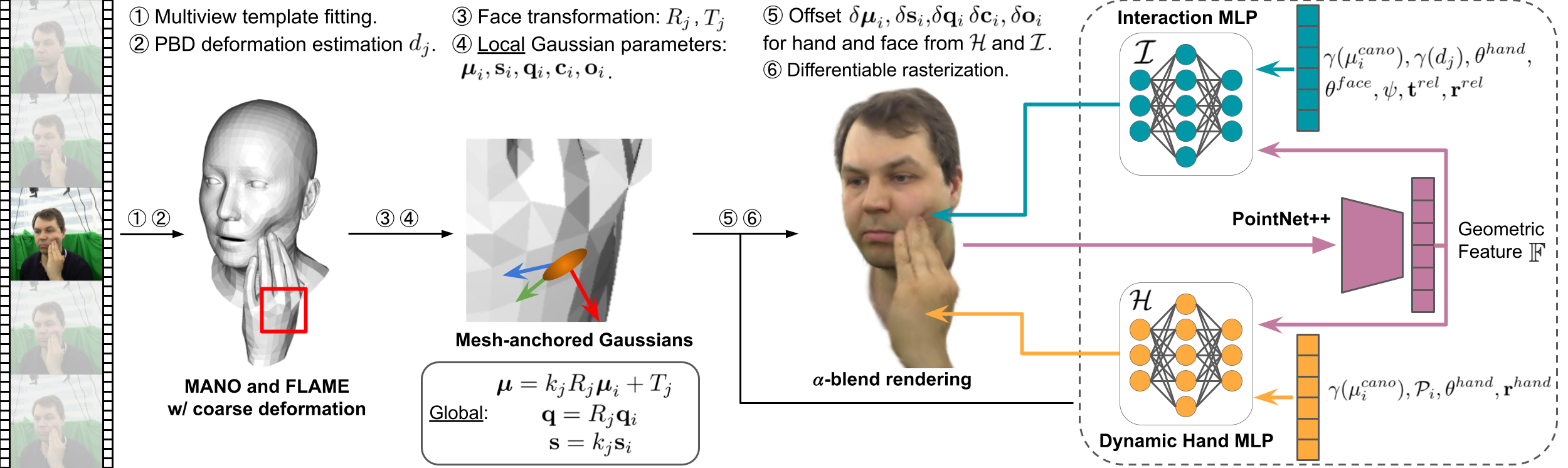}
    \caption{
    \textbf{Overview of \modelname}. Our method combines mesh-based geometry (FLAME, MANO) with 3D Gaussian Splatting for realistic hand-face interactions. The dynamic hand appearance module refines pose-dependent deformations, wrinkles, and shadows, while the Hand-Face Interaction module enhances contact-aware geometry and shading adjustments. This enables high-fidelity animation with lifelike interactions and appearance changes.}
    \label{fig:method}
\end{figure*}

\section{Method}
\label{sec:method}

\subsection{Preliminary}
3D Gaussian Splatting~\cite{kerbl3Dgaussians} adopts explicit anisotropic Gaussians to reconstruct static scenes or objects from multi-view images and known camera parameters. Each Gaussian is defined by the position $\mu$ and a covariance matrix $\Sigma$ representing shape:
\begin{equation}
G(\mathbf{x}) =  \exp\left(-\frac{1}{2}(\mathbf{x}-\boldsymbol{\mu})^T\Sigma^{-1}(\mathbf{x}-\boldsymbol{\mu})\right)
\end{equation}




We can use a volume rendering formula similar to NeRF~\cite{mildenhall2020nerf} to $\alpha$-blend the $N$ Gaussians and compute the color $C$ of a pixel:

\begin{equation}
C(\mathbf{x}) = \sum_{i=1}^{N} \mathbf{c}_i\alpha_i \prod_{j=1}^{i-1}(1-\alpha_j), \quad \alpha_i = o_i G_{i}'(\mathbf{x})
\end{equation}
where $c_i$ is the color of each Gaussian point parameterized by spherical harmonics. The blending weight $\alpha$ is evaluated based on the opacity parameter $o$ of the projected 2D Gaussians $G'$ sorted by depth order. 
\subsection{\modelname}
We propose a novel approach to modeling hand-face interactions using an explicit hybrid mesh-Gaussian representation with dynamic deformation and appearance modules.
Our method consists of: (1) a hybrid mesh-Gaussian avatar model for face and hand; (2) a dynamic hand appearance module that captures pose-dependent variations such as self-cast shadows and wrinkles during articulation; and (3) a hand-face interaction module that captures non-rigid deformation between these key body parts, as well as any deformation-induced changes in appearance. 

Our hybrid representation combines the strengths of both the mesh and Gaussian models. The mesh provides a coarse estimation of deformation, and the Gaussians on the surface provide photorealistic appearance and changes in fine details. Specifically, we use FLAME \cite{FLAME:SiggraphAsia2017} and MANO \cite{MANO:SIGGRAPHASIA:2017} as template mesh models that capture the shape, pose, and expression of the target human subject. We then build a layer of Gaussians on the mesh and bind the Gaussians to each mesh facet so that as the mesh moves and articulates, the Gaussians consistently follow, making it easy to animate and capture highly dynamic human subjects.

Our model is trained on multi-view videos of a subject performing hand-face interactions. At each time step, we get the FLAME and MANO mesh parameters for face and hand from multi-view template fitting~\cite{DecafTOG2023}, which requires multi-view observations and known camera calibration. We train the Gaussian avatar to reconstruct both hand and face, and use MLPs to model pose-dependent dynamic effects for the hand and interaction-induced deformation between hand and face.


\paragraph{Hybrid Mesh-Gaussian Face Avatar.} Our face Gaussians are combined with the FLAME mesh that parameterizes the facial mesh via the pose $\theta^{face}$, shape $\beta^{face}$, expression $\psi$, and global translation $\mathbf{t}^{face}$ and rotation $\mathbf{r}^{face}$. 

Following the definition in \cite{qian2024gaussianavatars}, for each mesh facet, we build a local coordinate frame based on the scale and orientation of the mesh. Each 3D Gaussian is bound to a mesh facet by assigning a face index. All spatial properties of the bound Gaussians are defined in the local triangle mesh and can freely move during optimization. 

Globally, the Gaussians can be animated and articulated along with the mesh as the local mesh coordinate basis rotates and translates. We construct the local coordinate system by setting the origin at the center of the triangle and form the orthogonal basis from the triangle normal and edge to obtain the rotation matrix $R$, which transforms the local Gaussians to the global world frame. We then define a scaling factor $k$ proportional to the size of the triangle so that the local position and scaling of Gaussians are defined relative to the scale of their parent triangle. This also ensures that during optimization 3D Gaussians have updates that are proportional to their parent triangle size. 

Specifically, for each Gaussian (indexed by $i$) on its parent triangle (indexed by $j$), we define the global transformation of the Gaussian as:
\begin{equation}
\boldsymbol{\mu} = k_jR_j\boldsymbol{\mu}_i+T_j
\label{eqn:mu_transform}
\end{equation}
\begin{equation}
\mathbf{s} = k_j\mathbf{s}_i
\label{eqn:s_transform}
\end{equation}
\begin{equation}
\mathbf{q} = R_j \mathbf{q}_i
\label{eqn:q_transform}
\end{equation}

During optimization, the number of Gaussians needs to be adjusted to correctly fit the topology, so we adopt adaptive density control \cite{kerbl3Dgaussians,qian2024gaussianavatars} to populate or prune the Gaussians based on the view-space positional gradient and opacity. To preserve the binding to the correct local mesh coordinates as we clone or split a Gaussian bound to a mesh face, we assign the same parent mesh index to the new Gaussian so that all new Gaussians are sampled close to the old ones for a smooth optimization process.

\paragraph{Hybrid Mesh-Gaussian Hand Avatar.} Our hand Gaussians are similarly combined with the MANO mesh that parameterizes the hand topology via the estimated pose $\theta^{hand}$, shape $\beta^{hand}$, and global translation $\mathbf{t}^{hand}$ and rotation $\mathbf{r}^{hand}$. Note that, to the best of our knowledge, this is the first avatar work to obtain a hybrid representation of hands by anchoring Gaussian kernels to mesh. Previous work like MANUS~\cite{pokhariya2024manus} anchors Gaussian kernels to a skeleton and learns inverse skinning weights but not to the mesh.

As described above for our face Gaussians, we follow the same procedure to build a layer of Gaussians to represent the hand on the MANO mesh and use the transformations outlined in Eqs. \ref{eqn:mu_transform}-\ref{eqn:q_transform} to obtain the global position, scale, and rotation of the Gaussians when animating the hand.


\paragraph{Dynamic Hand Appearance Refinement.} 
Compared to the face, hands present a unique challenge due to complex articulations and visual dynamics, such as self-cast shadows, wrinkles, and intricate skin deformations. Static Gaussian parameters are therefore not sufficient to capture high-fidelity details for hands. To address this gap, we model geometry and appearance dynamics using a set of MLPs to estimate offset on each Gaussian parameter:

\begin{equation}
\delta\boldsymbol{\mu}_i, \delta\mathbf{s}_i, \delta\mathbf{q}_i\ = \mathcal{H}_{geo}(\gamma(\mathbf{\mu}_i^{cano}), \theta^{hand})
\end{equation}

\begin{equation}
\delta\mathbf{c}_i, \delta\mathbf{o}_i\ = \mathcal{H}_{app}(\gamma(\mathbf{\mu}_i^{cano}),\mathcal{P}_i, \mathbb{F}, \theta^{hand}, \mathbf{r}^{hand}, \mathbf{r}^{rel}, \mathbf{t}^{rel}) 
\end{equation}
where $\gamma(\mathbf{\mu}_i^{cano})$ is the positional encoding of the canonical position of the Gaussian  (set by the first frame of the training sequence), $\mathcal{P}_i\in \mathbb{R}^{64}$ is the per point feature associated with each Gaussian that is optimized together with the MLPs, $\mathbb{F}$ encodes the geometric features which is detailed in the next section, while $\mathbf{r}^{rel}$ and $\mathbf{t}^{rel}$ are the relative orientation and translation between the hand and face which models interaction-induced effects on hand.  

The geometry MLP $\mathcal{H}_{geo}$ captures the intricate pose-dependent shape dynamics represented by the relative change in position ($\delta\mathbf{\mu}_i$), scale ($\delta\mathbf{s}_i$), and rotation ($\delta\mathbf{q}_i$) of each Gaussian, which are added to the corresponding variables in Eqs. \ref{eqn:mu_transform}-\ref{eqn:q_transform}.

The appearance MLP $\mathcal{H}_{app}$ uses an auto-decoder to capture the appearance change (\ie relative change in color $\delta\mathbf{c}_i$ and opacity $\delta\mathbf{o}_i$) dependent on the hand pose $\theta^{hand}$ and orientation $\mathbf{r}^{hand}$. We further incorporate a point feature embedding $\mathcal{P}_i \in \mathcal{R}^{64}$ as an additional parameter for each Gaussian that is jointly optimized with the MLP. We use $\mathcal{H}_{app}$ to update color and opacity as $\mathbf{c}_i=\mathbf{c}_i+\delta\mathbf{c}_i$ and $\mathbf{o}_i=\mathbf{o}_i+\delta\mathbf{o}_i$, respectively.

The relative changes in each parameter are then added to obtain the resultant Gaussian parameters:

\begin{equation}
\mathbf{\mu} = k_jR_j(\mathbf{\mu}_i + \delta\mathbf{\mu}_i)+T_j
\label{eqn:mu_delta_transform}
\end{equation}
\begin{equation}
\mathbf{s} = k_j(\mathbf{s}_i + \delta\mathbf{s}_i)
\label{eqn:s_delta_transform}
\end{equation}
\begin{equation}
\mathbf{q} = R_j (\mathbf{q}_i + \delta\mathbf{q}_i)
\label{eqn:q_delta_transform}
\end{equation}
\begin{equation}
    \mathbf{c}_i = \mathbf{c}_i + \delta\mathbf{c}_i
    \label{eqn:c_delta}
\end{equation}
\begin{equation}
    \mathbf{o}_i = \mathbf{o}_i + \delta\mathbf{o}_i
    \label{eqn:o_delta}
\end{equation}

\paragraph{Hand-Face Interaction.}
To simulate the deformation changes due to hand and face interactions, we can leverage a physical simulation method like position-based dynamics (PBD) \cite{Mller2007PositionBD}, which avoids overfitting and generalization challenges when learning physical deformations from data. We follow DECAF \cite{DecafTOG2023} to resolve hand-face collisions and obtain per-vertex 3D offsets of the facial geometry of interacting hand-face deformation. To obtain more anatomically realistic deformation, DECAF \cite{DecafTOG2023} computes the stiffness value for each face vertex based on skin-skull distance. As the facial mesh deforms, the local Gaussians also deform along with the mesh. However, the coarsely estimated deformation may present a gap between the predicted and actual deformations. To address this deformation gap as well as appearance changes (e.g., shadows), we propose a set of interaction MLPs:

\begin{equation}
\begin{split}
(\delta\boldsymbol{\mu}_i, \delta\mathbf{s}_i, & \delta\mathbf{q}_i\, \delta\mathbf{c}_i, \delta\mathbf{o}_i) = \\
&\mathcal{I}(\gamma(\mathbf{\mu}_i^{cano}), \gamma(d_j), \mathbb{F}, \theta^{hand}, \theta^{face}, \psi, \mathbf{t}^{rel}, \mathbf{r}^{rel})
\end{split}
\end{equation}
where, similar to the dynamic Gaussian hand, the deformation dynamics are modeled as relative change in position, scale, and rotation for each Gaussian and appearance dynamics are modeled as the relative change in color and opacity. The hand-face interaction MLPs $\mathcal{I}$ take as input the positional encoding of the canonical position $\gamma(\mathbf{\mu}_i^{cano})$ and deformation offset from the parent mesh face $\gamma(d_j)$, the global geometric feature $\mathbb{F}$, the pose of the hand $\theta^{hand}$ and face $\theta^{face}$, facial expression $\psi$ (estimated from FLAME), as well as the relative translation $\mathbf{t}^{rel}$ and rotation $\mathbf{r}^{rel}$ between the hand and face. To compute the deformation offset $d_j$, we take the average offset of the parent mesh face vertices $d_j = \frac{1}{3}\sum(V_1+V_2+V_3)$. 

In order to capture the local and global relationships among hand and face Gaussians, we use PointNet++~\cite{qi2017pointnetplusplus} to extract the geometric feature $\mathbb{F}\in\mathbb{R}^{1024}$. However, it is computationally infeasible to consider hundreds of thousands of Gaussians in this manner, so we instead propose an efficient sampling heuristic. First, we consider only the Gaussians located at non-rigid facial regions where deformation is possible. Next, we sample a single Gaussian from each mesh facet proportional to $\mathbf{o}_i ||\mathbf{s}_i||_2$, since the Gaussian with highest opacity and largest scale contributes most to the local geometry.

Since some regions of the face are more non-rigid than others (\eg cheek), we can focus updates within these regions to improve computational efficiency by selecting only the mesh faces that have a deformation greater than a minimal threshold. Intuitively, larger deformations occur in regions closest to the contact point between the hand and face. We therefore apply weights $w_i$ to the predicted offsets inversely proportional to the distance between the Gaussian and nearest hand vertex, as shown in the following equations.

\begin{equation}
w_i = \frac{1}{2}(cos(\pi\frac{\min_{j} \|\boldsymbol{\mu}_i - \boldsymbol{V}_j\|_2}{d_{max}}) + 1)
\end{equation}

\begin{equation}
\mathbf{x}_i = \mathbf{x}_i + w_i\delta \mathbf{x}_i, \quad \mathbf{x}_i=\{\boldsymbol{\mu}_i, \mathbf{s}_i, \mathbf{q}_i, \mathbf{c}_i, \mathbf{o}_i\}
\end{equation}
where $d_{max}=0.05$ is the maximum distance and anything exceeds this limit has $w_i=0$.

\subsection{Training and Regularization}
Our method follows a two-stage training paradigm with an initial warm-up phase to train the static Gaussian parameters followed by training of the hand and interaction MLPs $\mathcal{H}$ and $\mathcal{I}$.
In the first stage, we set the first frame in the training sequence as the canonical frame and initialize $N$ Gaussians at the center of each face on the hand and face meshes. We initialize the rest of the Gausssian parameters as done in \cite{kerbl3Dgaussians}. We supervise the rendered images with $\mathcal{L}_1$ and SSIM losses:

\begin{equation}
\mathcal{L} = (1-\lambda)\mathcal{L}_1 + \lambda\mathcal{L}_{D-SSIM}
\end{equation}

During the second stage of training, we initialize the weights of the final layers of the hand and interaction MLPs with zeros in order to fine-tune intricate details without negatively impacting the overall structure early in training. 

When training the hand and interaction MLPs, we identify a few challenges that must be addressed. First, since hand-face interactions are relatively rare occurrences within video sequences, we oversample video frames where such interactions do occur to ensure sufficient training data. Second, when animating the Gaussian avatar to novel poses, we occasionally encounter thin and spiky Gaussians, which are exacerbated on the hand by significant articulations of the fingers. Intuitively, larger Gaussians are more likely to contribute to such visual artifacts because a small rotation could be greatly magnified by the scale. To mitigate this, we regularize the scaling to encourage smaller and more isotropic Gaussians:
\begin{equation}
\mathcal{L}_{s} = \|ReLU(\mathbf{s} + \delta\mathbf{s} - \epsilon_{s}) \|_2^2
\end{equation}
where $\epsilon_{s}$ is a minimal threshold for the scaling parameter to prevent the Gaussians from shrinking excessively.

Third, Gaussians can deviate from the coarse topology of the mesh in an unrealistic manner. To prevent such deviation from the parent mesh, we further add a position regularization loss:

\begin{equation}
\mathcal{L}_{\mu} = \|ReLU(\mathbf{\mu} + \delta\mathbf{\mu} - \epsilon_{\mu}) \|_2^2
\end{equation}
where $\epsilon_{\mu}$ defines the area around the mesh that Gaussians are free to move about in order to adjust the discrepancies between the fitted mesh and visual observation. We only apply $\mathcal{L}_{s}$ and $\mathcal{L}_{\mu}$ to visible Gaussians to preserve the structure of occluded parts as the hand articulates and interacts with the face.

Finally, since hands and hand-face interactions usually comprise a small area of the scene yet have rich visual details, we track the hand and face with bounding boxes and add a patch perceptual loss $\mathcal{L}_{patch}$ \cite{zhang2018perceptual} that targets the hand as well as the overlapping region of hand and face bounding boxes. Our overall training loss is defined as follows:
\begin{equation}
\mathcal{L} = (1-\lambda)\mathcal{L}_1 + \lambda\mathcal{L}_{D-SSIM} + a\mathcal{L}_{s} + b\mathcal{L}_{\mu} + c\mathcal{L}_{patch}
\end{equation}

\section{Experiment}
\label{sec:experiment}
\subsection{Setup}
\paragraph{Implementation details.} 
We follow~\cite{DecafTOG2023} to obtain MANO and FLAME mesh parameters and coarse mesh deformation from multiview videos by fitting 2D projected keypoints given known camera calibrations and solving PBD optimization.
We initialize 20 Gaussians per mesh face on the mesh, assume no view-dependent effects, and disable spherical harmonics. We use 4-layer and 6-layer MLPs with 256 hidden units and leaky ReLU activation, layer norm, and a skip connection from the input to the middle layer for the dynamic hand module $\mathcal{H}$ and interaction module $\mathcal{I}$ respectively. We designate an MLP for each Gaussian parameter. 
We first train the static Gaussian face and hand avatars for $100k$ steps as a warm-up before training the hand and hand-face interaction MLPs.
We use the position embedding in~\cite{mildenhall2020nerf} to encode Gaussian positions. 

We use Adam~\cite{kingma2017adammethodstochasticoptimization} to optimize our model and use the same learning rate and exponential learning decay as in the original 3D Gaussian Splatting~\cite{kerbl3Dgaussians}. We set the learning rate for MLPs to $1e-3$ and the learning rate for the point feature $2.5e-3$. We set loss weights $\lambda=0.2$, $a=1.0$, $b=0.01$, $c=0.1$ and the threshold $\epsilon_s=0.4$ and $\epsilon_\mu=0.2$. We train \modelname on multi-view videos using an NVIDIA RTX 3090 GPU. 

\paragraph{Datasets and evaluation.}
We evaluate novel view reconstruction, self-enactment on novel sequences, and cross-actor enactment of hand-face interaction using the only public available 3D hand-face interaction dataset DECAF~\cite{DecafTOG2023}. We take four subjects and use seven views to train while holding out one view to test novel views. We use the train and test split in DECAF for self-enactment and cross-actor enactment evaluation. We use SAM 2~\cite{ravi2024sam2} to segment the hand and face  then crop and resize the images to $512\times512$ using the tracked hand and face bounding boxes. For quantitative metrics, we report PSNR, SSIM, and LPIPS.

\paragraph{Baselines}We select recent state-of-the-art Gaussian face avatars like GaussianAvatar~\cite{qian2024gaussianavatars} and SplattingAvatar~\cite{shao2024splattingavatar} as our baselines. They are both the latest hybrid Gaussian head avatars anchored on a template face or body mesh. We naively extend their representation to modeling hands driven by MANO hand template mesh in our setting. 

\subsection{Dynamic Gaussian Hand}
\begin{figure}[t]
    \centering
    \includegraphics[width=0.9\columnwidth]{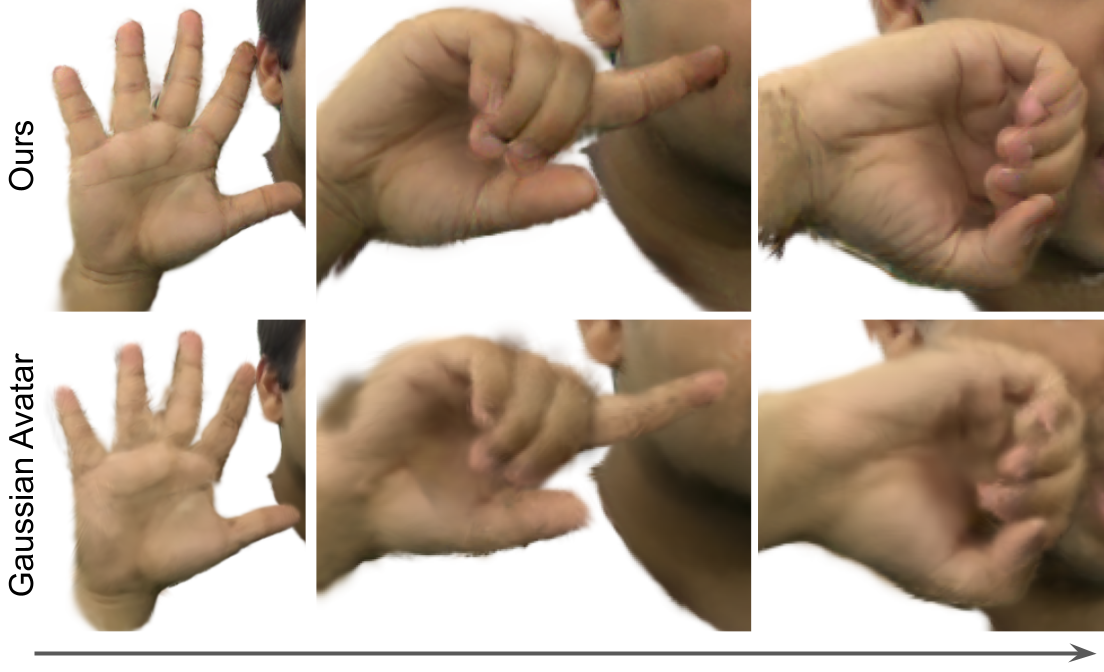}
    \caption{
    \textbf{Dynamic Gaussian Hand} adapts to pose, capturing self-cast shadows, wrinkles, and shading variations. The baseline methods struggle with static hand modeling, whereas our approach preserves fine-grained details across diverse hand poses.}
    \label{fig:dynamic_hand}
\end{figure}
\begin{figure*}[!tp]
    \centering
    \includegraphics[width=\textwidth]{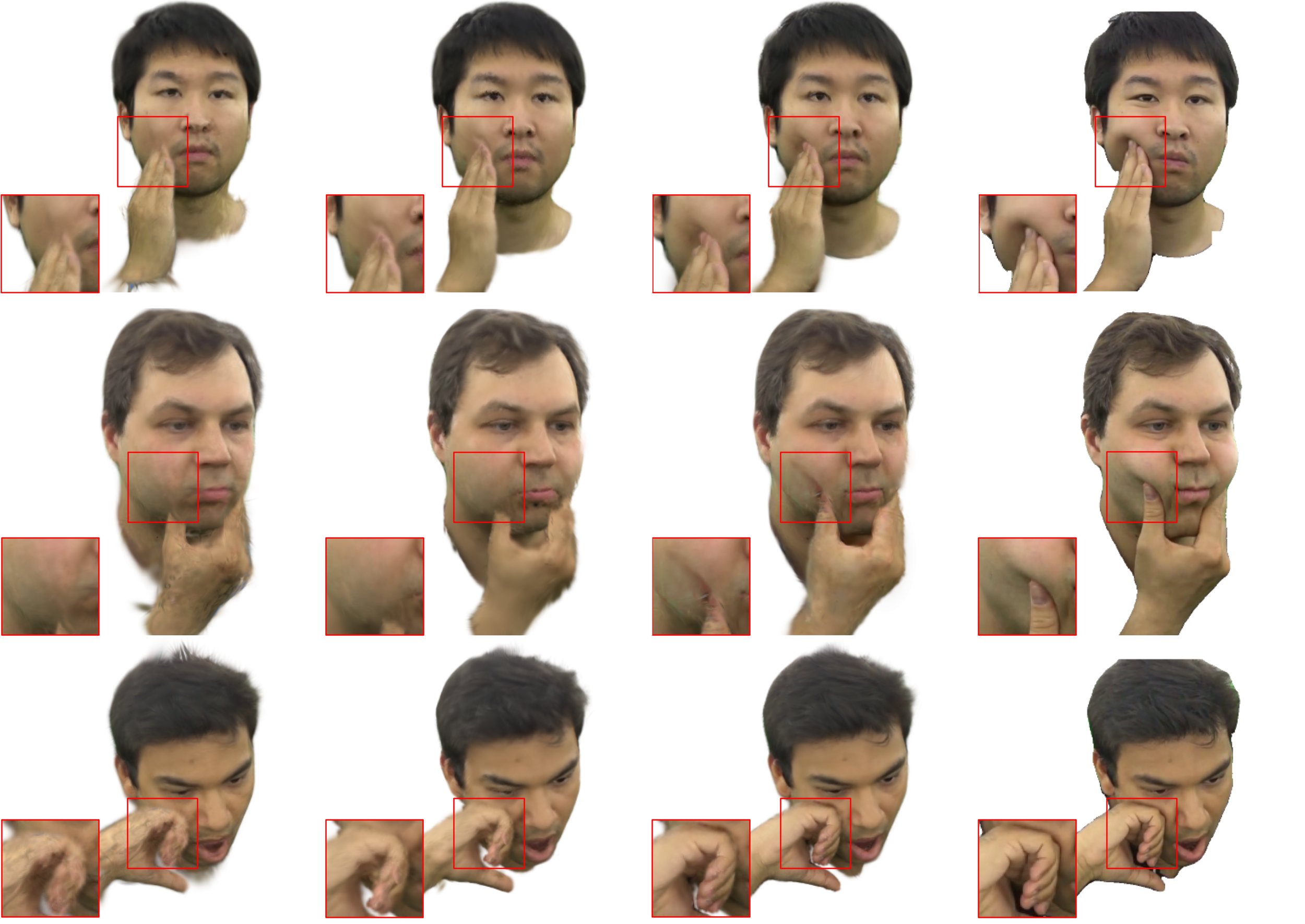}\\
    \makebox[\textwidth]{%
        \makebox[0.25\textwidth][c]{Splatting Avatar~\cite{shao2024splattingavatar}}%
        \makebox[0.25\textwidth][c]{Gaussian Avatar~\cite{qian2024gaussianavatars}}%
        \makebox[0.25\textwidth][c]{Ours}%
        \makebox[0.25\textwidth][c]{Ground Truth}%
    }
    \caption{
    \textbf{Qualitative Comparison of Hand-Face Interactions from Novel Views}. Our method produces sharp, high-fidelity details on non-rigid facial deformations and dynamic hand appearances, outperforming baseline models like GaussianAvatar~\cite{qian2024gaussianavatars} and SplattingAvatar~\cite{shao2024splattingavatar} Features like shadowing, wrinkles, and natural hand-face deformations are accurately reconstructed.}
    \label{fig:nvs}
    \vspace{-0.2in}
\end{figure*}

In Figure~\ref{fig:dynamic_hand} and bottom row in Figure~\ref{fig:nvs}, we demonstrate that our proposed hand module captures high-fidelity details of pose-dependent dynamics such as self-cast shadows and wrinkles. The baseline GaussianAvatar and SplattingAvatar only allow Gaussian geometry to morph based on the transformation underlying mesh they are bounded to. This limits the exprerssiveness of the Gaussian geometry to capture highly articulated hands and assumes static appearance as color and opacity remain the same. Thanks to the dynamic hand MLPs $\mathcal{H}$, we can learn the change in geometry and appearance dependent on hand pose and orientation.

\begin{table}[!tp]
    \centering \footnotesize
    \begin{tabular}{ccccc}
        \hline
        && PSNR$\uparrow$ & SSIM$\uparrow$ & LPIPS$\downarrow$   \\
        \hline
        &SplattingAvat.~\cite{shao2024splattingavatar} &  25.90& 0.9135 & 0.0584  \\
        Novel view&GaussianAvat.~\cite{qian2024gaussianavatars} & 26.71 & 0.9324 & 0.0542  \\
        &Ours & \textbf{29.85} & \textbf{0.9335} & \textbf{0.0338}    \\
        \hline
        &SplattingAvat.~\cite{shao2024splattingavatar} & 26.18 & 0.9326 & 0.0546 \\
        Self-enact&GaussianAvat.~\cite{qian2024gaussianavatars} & 25.51 & \textbf{0.9393} & 0.0578  \\
        &Ours &  \textbf{28.17}& 0.9356 &  \textbf{0.0396}  \\ 
        \hline
    \end{tabular}
    \caption{
    \textbf{\modelname outperforms baselines quantitvely} on PSNR and LPIPS across novel view synthesis and self-enactment.}
    \label{tab:results}
\end{table}
\begin{table}[!tp]
    \centering \footnotesize
    \begin{tabular}{cccccc}
        \hline
        && PSNR$\uparrow$ & SSIM$\uparrow$ & LPIPS$\downarrow$  \\
        \hline
        Ours &&   \textbf{29.85} & 0.9335 & \textbf{0.0338}  \\
        w/o hand MLPs $\mathcal{H}$ &&  26.97 & 0.9359 & 0.0574   \\
        w/o interaction MLPs $\mathcal{I}$ &&  28.23 & \textbf{0.9389} &  0.0467  \\
        \hline
    \end{tabular}
    \caption{
    \textbf{Ablation Study on Novel View Synthesis}. Our full model provides the best perceptual quality.}
    \label{tab:ablation}
    \vspace{-0.2in}
\end{table}






\begin{figure*}[!tp]
    \centering
    \includegraphics[width=\textwidth]{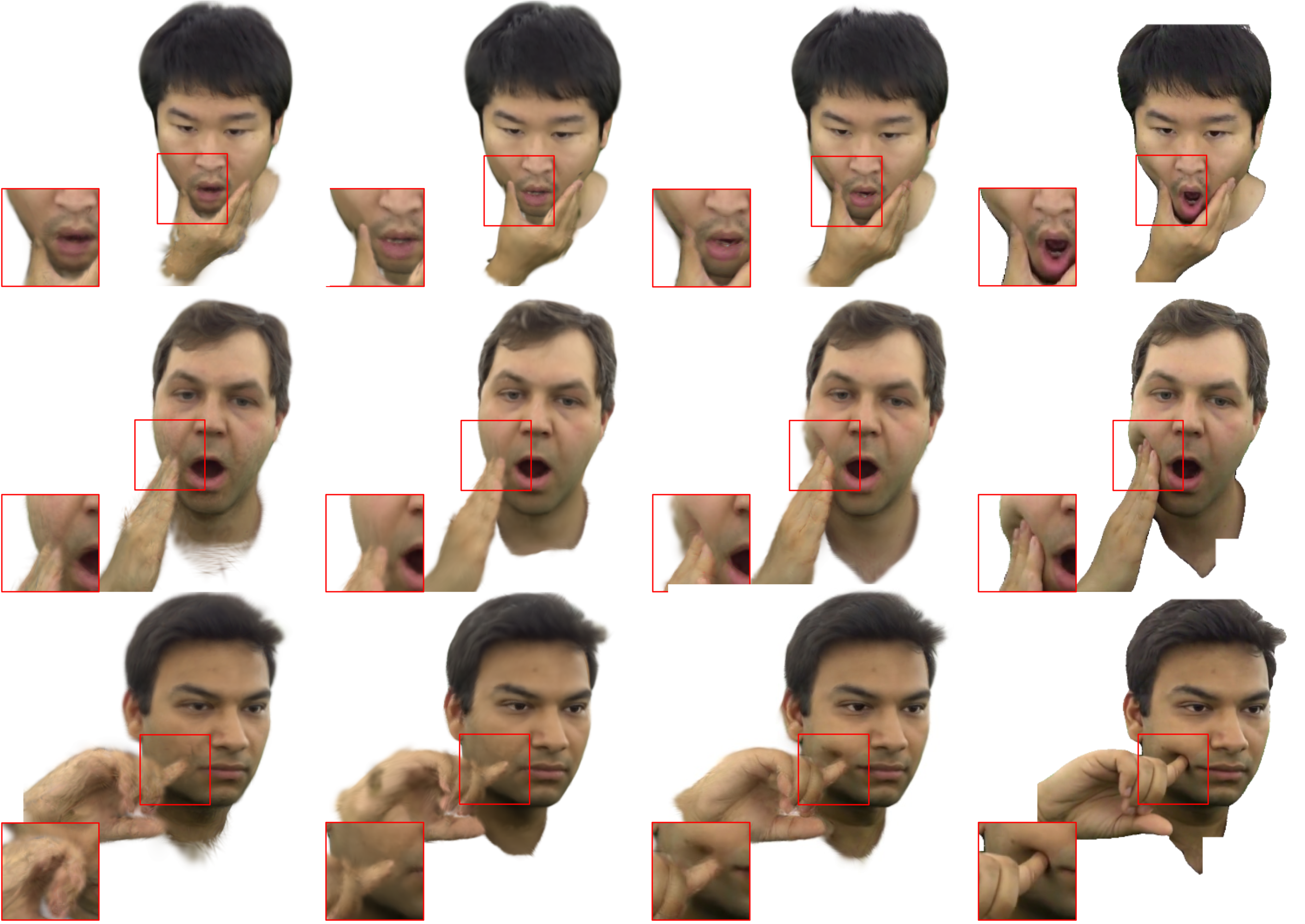}\\
    \makebox[\textwidth]{%
        \makebox[0.25\textwidth][c]{Splatting Avatar~\cite{shao2024splattingavatar}}%
        \makebox[0.25\textwidth][c]{Gaussian Avatar~\cite{qian2024gaussianavatars}}%
        \makebox[0.25\textwidth][c]{Ours}%
        \makebox[0.25\textwidth][c]{Ground Truth}%
    }
    \caption{
    \textbf{Self-Enactment with \modelname}. Our method accurately reconstructs natural hand-face interactions, preserving fine geometric details and appearance consistency in self-enactment tasks. Compared to baselines, \modelname effectively models dynamic wrinkles, shadows, and subtle hand-induced deformations.
    }
    \label{fig:self_enactment}
    \vspace{-0.2in}
\end{figure*}
\begin{figure}[t]
    \centering
    \includegraphics[width=\columnwidth]{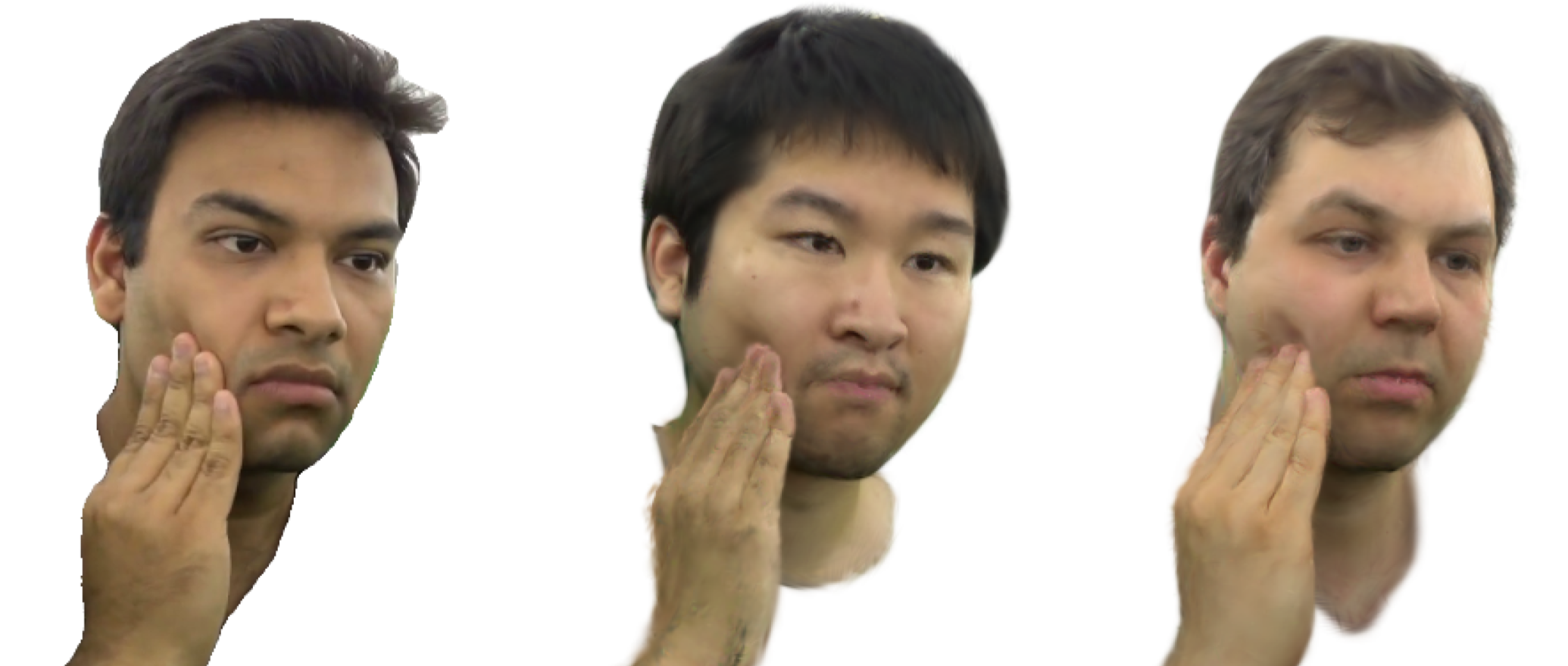}
    \makebox[\columnwidth]{%
        \makebox[0.33\columnwidth][c]{Source actor}%
        \makebox[0.66\columnwidth][c]{Target avatars}
    }
    \caption{
    \textbf{Cross-Actor Re-enactment}. We use the pose $\theta^{face}$, $\theta^{hand}$ and expression $\psi$ parameters from the tracked FLAME and MANO templates of the source actor to drive the target avatars. Our \modelname can generalize to novel motions and interaction-induced dynamics.}
    \label{fig:cross_actor_enactment}
    \vspace{-0.2in}
\end{figure}

\subsection{Hand-Face Interaction}
\begin{figure}[t]
    \centering
    \includegraphics[width=\columnwidth]{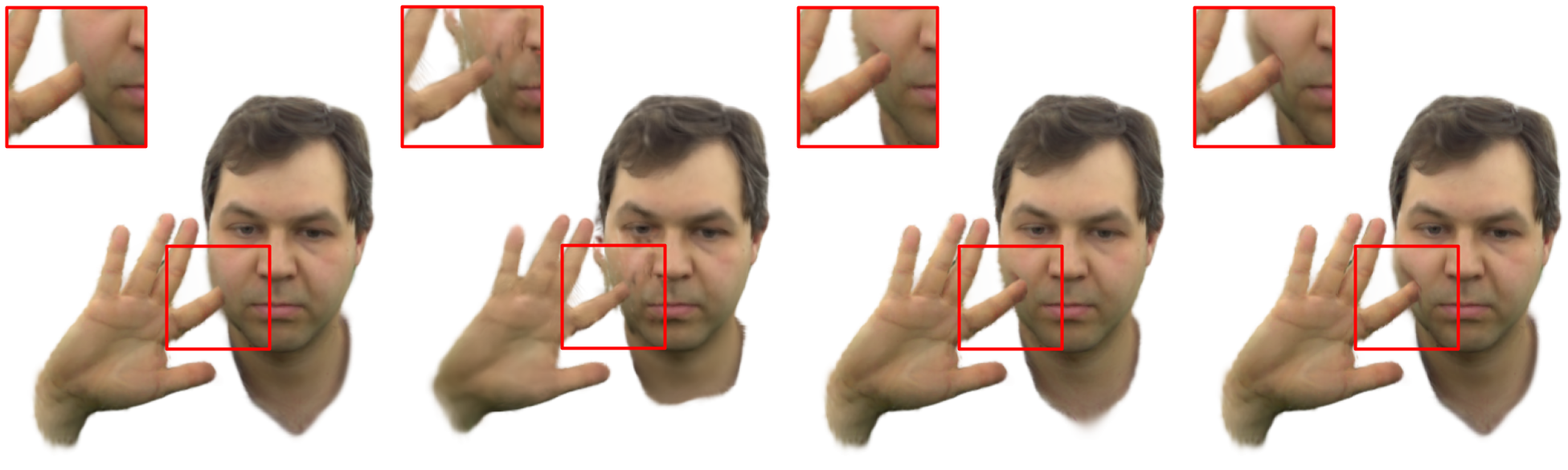}
    \makebox[\columnwidth]{%
        \makebox[0.25\columnwidth][c]{w/o PBD }%
        \makebox[0.25\columnwidth][c]{w/o MLPs $\mathcal{I}$}%
        \makebox[0.25\columnwidth][c]{w/o $\mathcal{L}_{patch}$}%
        \makebox[0.25\columnwidth][c]{Ours}%
    }
    \caption{
    \textbf{Ablation}. Red highlights: (1) PBD provides coarse geometric deformation on mesh. (2) Naively training with PBD deformation without the interact MLPs $\mathcal{I}$ to compensate the geometry would result in "flying out" Gaussians.
    (3) The interaction regions are small in the image space and provides weak training signal, we add a patch perceptual loss $\mathcal{L}_{patch}$ to better capture the fine visual details. }
    \label{fig:ablation_fig}
    \vspace{-0.2in}
\end{figure}

We evaluate the quality of reconstruction on held-out novel views (Figure~\ref{fig:nvs}) and animation quality on self-reenactment (Figure~\ref{fig:self_enactment}) using hand and head poses in test sequence to drive the trained avatar. We show quantitative evaluation in Table~\ref{tab:results}. While PSNR and LPIPS are greatly improved, SSIM in our self-enactment is slightly worse than Gaussian Avatar, because our major improvements focus on sharp dynamic details on hand and interaction region but the MLPs may introduce some noises in the global structure of face and hand in test sequences due to generalization gap. Both Splatting Avatar and Gaussian Avatar show great quality in face reconstruction and animation. However, they struggle to reconstruct hands due to their complex articulation and dynamic appearance compared with facial expressions. Splatting Avatar allows Gaussians to walk between the mesh faces yet demonstrate worse quality than Gaussian Avatar which bounds the Gaussians to particular mesh faces. Gaussian Avatar shows over-smoothed appearance on hands as it doesn't model dynamic color and opacity as the geometry changes. Both baselines fail to capture the non-rigid deformation and appearance dynamics when hand interacts with face. Our method captures the complex dynamics of both hand and hand-face interaction by encoding pose and geometry with MLPs. In Figure~\ref{fig:cross_actor_enactment}, we demonstrate that our interaction module can be generalized to motions from different actors, showcasing applications in film, gaming, and VR. 
We did ablation studies on subject $\#4$ and report results in Table~\ref{tab:ablation} and Figure~\ref{fig:ablation_fig}, demonstrating the effects of each design choice. Particularly, without our interaction module, naively training with PBD mesh deformation only produces artifacts in Gaussian geometry shown in the 2nd example of Figure~\ref{fig:ablation_fig}.



\subsection{Limitations}
Despite some promising results, it is difficult to handle the vast complexity of hand-face interactions. While the dataset we use captures some common behaviors, it cannot exhaustively cover the countless possible configurations of hand-face dynamics, potentially limiting generalization to novel scenarios in the wild. Given the scarcity of specialized datasets for human self-interaction, we hope our research will bring greater attention to this important yet underexplored problem within the research community.
\section{Conclusion}
We address a valuable and challenging problem for entertainment and social media industries: hand-face interaction for human avatars. We present \modelname, the first method to model dynamic hands and hand-face interactions. We do so by proposing a novel hybrid mesh-Gaussian representation of hands that models deformations and appearance dynamics during articulation. We further propose a novel hand-face interaction module that models geometry and appearance changes during hand-face interaction. Our technique can model realistic deformation, render complex details like shadows, generalize to unseen poses, and enable reenactment. Future work will focus on improving generalization for complex hand-face interactions.
\label{sec:conclusion}
{
    \small
    \bibliographystyle{ieeenat_fullname}
    \bibliography{main}
}

\end{document}